%% file: templateArxiv.tex
\documentclass{article}

\usepackage{PRIMEarxiv}

\usepackage[utf8]{inputenc} 
\usepackage[T1]{fontenc}    
\usepackage{hyperref}       
\usepackage{url}            
\usepackage{booktabs}       
\usepackage{amsfonts}       
\usepackage{nicefrac}       
\usepackage{microtype}      
\usepackage{lipsum}
\usepackage{fancyhdr}       
\usepackage{graphicx}       
\graphicspath{{media/}}     
\usepackage{fontawesome}
\usepackage{subfigure}
\usepackage{multirow}
\usepackage{multicol}
\usepackage{xcolor}
\usepackage{amsmath}
\newif\ifshowcomment
\showcommenttrue

\ifshowcomment
\newcommand{\yuanye}[1]{\textcolor{orange}{[Yuanye: #1]}}
\newcommand{\todo}[1]{\textcolor{red}{[TODO: #1]}}
\else
\newcommand{\yuanye}[1]{}
\newcommand{\todo}[1]{}
\fi

\title{AquilaMoE: Efficient Training for MoE Models with Scale-Up and Scale-Out Strategies
}

\author{
  \textbf{Bo-Wen Zhang}, \textbf{Liangdong Wang}, \textbf{Ye Yuan}, \textbf{Jijie Li}, \textbf{Shuhao Gu}, \textbf{Mengdi Zhao}, \textbf{Xinya Wu}, \textbf{Guang Liu}\thanks{Project Lead, the corresponding author, contact \url{liuguang@baai.ac.cn}}, \\
  \textbf{Chengwei Wu}, \textbf{Hanyu Zhao}, \textbf{Li Du}, \textbf{Yiming Ju}, \textbf{Quanyue Ma}, \textbf{Yulong Ao}, \textbf{Yingli Zhao}, \textbf{Songhe Zhu}, \textbf{Zhou Cao}, \\
  \textbf{Dong Liang}, \textbf{Yonghua Lin}, \textbf{Ming Zhang}, \textbf{Shunfei Wang}, \textbf{Yanxin Zhou}, \textbf{Min Ye}, \textbf{Xuekai Chen}, \textbf{Xinyang Yu}, \\
  \textbf{Xiangjun Huang}, \textbf{Jian Yang}\thanks{Full authorship contribution statements appear at the end of the document.} \\
    Beijing Academy of Artificial Intelligence (BAAI) \\
    School of Computer Science, Peking University \\
    MetaX-Tech\\
}

\begin{document}
\maketitle

\begin{abstract}
In recent years, with the rapid application of large language models across various fields, the scale of these models has gradually increased, and the resources required for their pre-training have grown exponentially. Training an LLM from scratch will cost a lot of computation resources, while scaling up from a smaller model is a more efficient approach and has thus attracted significant attention. In this paper, we present AquilaMoE, a cutting-edge bilingual 8*16B Mixture of Experts (MoE) language model that has 8 experts with 16 billion parameters each and is developed using an innovative training methodology called \textbf{EfficientScale}. This approach optimizes performance while minimizing data requirements through a two-stage process. The first stage, termed \textit{Scale-Up}, initializes the larger model with weights from a pre-trained smaller model, enabling substantial knowledge transfer and continuous pretraining with significantly less data. The second stage, \textit{Scale-Out}, uses a pre-trained dense model to initialize the MoE experts, further enhancing knowledge transfer and performance. Extensive validation experiments on 1.8B and 7B models compared various initialization schemes, achieving models that maintain and reduce loss during continuous pretraining. Utilizing the optimal scheme, we successfully trained a 16B model and subsequently the 8*16B AquilaMoE model, demonstrating significant improvements in performance and training efficiency.
\end{abstract}


\section{Introduction}

Language models have become a cornerstone of modern natural language processing (NLP) systems, driving applications such as machine translation, conversational agents, text summarization, and question answering \cite{vaswani2017attention, devlin2018bert}. Recent advancements in large language models (LLMs) like GPT-3, BERT, and T5 have demonstrated remarkable proficiency across numerous tasks, highlighting the importance of pretraining on large-scale datasets to achieve state-of-the-art results \cite{brown2020language, raffel2020exploring}. Despite their success, traditional dense models face significant challenges in scalability and efficiency, particularly as parameter sizes increase.

Mixture of Experts (MoE) models have emerged as a promising solution to these challenges. By dynamically selecting different subsets of model parameters (experts) for various inputs, MoE architectures can scale to a much larger number of parameters without a corresponding increase in computational cost \cite{lepikhin2021gshard}. This selective activation mechanism allows MoE models to achieve higher performance while maintaining computational efficiency. However, training such large-scale MoE models presents significant challenges, including the vast amounts of data and computational power required.

Training large-scale models, including MoE architectures, involves several critical challenges. Traditional training methods require enormous amounts of data, which can be resource-intensive and time-consuming to collect and process. The computational cost is substantial, requiring high-performance hardware such as GPUs or TPUs, and significant energy consumption, making it challenging for many institutions with limited resources to train and deploy such models. Additionally, training large models from scratch can take weeks or even months, delaying experimentation and iteration. Ensuring that the model efficiently learns and generalizes well is also challenging, as poor initialization and inefficient training strategies can lead to suboptimal performance and wasted resources.

Several strategies have been proposed to address these challenges. For instance, the Net2Net method accelerates learning via knowledge transfer, allowing the seamless transition of knowledge from smaller to larger networks, which shows significant acceleration in image classification task~\cite{net2net}. The StackBERT method improves training efficiency by progressively increasing model depth and capacity \cite{stacking}. The bert2BERT approach focuses on reusing pre-trained language models to initialize new models, promoting efficiency and reusability~\cite{chen-etal-2022-bert2bert}. It expands both the width and depth of the smaller model and finally saves nearly half of the pre-training consumption of language models. The primary motivation behind developing AquilaMoE is to introduce an efficient training framework, EfficientScale, which reduces data and computational requirements while enhancing overall model performance. Our approach leverages the strengths of MoE architectures and introduces innovative techniques to improve training efficiency and effectiveness.

In this paper, we introduce AquilaMoE, a bilingual 8*16B Mixture of Experts language model that has 8 experts with 16 billion parameters each and is developed using the \textbf{EfficientScale} methodology. This approach optimizes performance and minimizes data needs through a two-stage process. The first stage, \textit{Scale-Up}, leverages the weights of a pre-trained smaller model to initialize the larger model, enabling substantial knowledge transfer and continuous pretraining with significantly less data compared to traditional from-scratch training. The second stage, \textit{Scale-Out}, uses a pre-trained dense model to initialize the MoE experts, further enhancing knowledge transfer and performance.

Through extensive validation experiments on 1.8B and 7B models, we compared various initialization schemes to achieve models that maintain and further reduce loss during continuous pretraining. Based on these findings, we utilized the optimal initialization scheme to successfully train a 16B model and subsequently the 8*16B AquilaMoE model, demonstrating significant advancements in model performance and training efficiency.

\section{Methodology}

The EfficientScale pipeline is designed to efficiently train a large-scale Mixture of Experts (MoE) model by leveraging knowledge transfer from smaller models. The process involves three main phases: Preparation, Scale-Up, and Scale-Out. Each phase plays a crucial role in ensuring effective knowledge transfer and continuous learning, resulting in a highly optimized MoE model.

\subsection{Preparation Phase}
The preparation phase involves training a small dense model and preparing the datasets required for subsequent phases. This phase ensures that the initial model has sufficient transferable knowledge and that the data is ready for effective training and validation.

\begin{itemize}
    \item \textbf{Model Training}: Train a small dense model from scratch on a substantial amount of tokens or use an already pre-trained small model. This step ensures the model has accumulated sufficient transferable knowledge to serve as a robust starting point.
    \item \textbf{Data Preparation}: Collect, clean, and preprocess the training and validation datasets. This step involves managing large datasets to ensure they are suitable for training and validation purposes.
    \item \textbf{Validation Setup}: Develop both training and validation datasets to monitor the model’s performance during subsequent phases. Continuous tracking of the language model’s loss on the validation dataset is essential to ensure the initialized models retain transferred knowledge and can learn new information effectively.
\end{itemize}

\input{sections/scale-up}

\subsection{Scale-Out Phase}
The scale-out phase involves transforming the large dense model into a Mixture of Experts (MoE) model. This phase includes initializing the MoE model's weights and performing continuous pretraining to refine the model's knowledge and performance.

\begin{itemize}
    \item \textbf{MoE Weight Initialization}: Aquila-MoE is initialized using Sparse Upcycling \cite{komatsuzaki2022sparse, hu10minicpm}. The dense model checkpoint obtained from the Aquila dense model undergoes a transformation where each MLP layer is replaced by an MoE layer. These new MoE layers are exact replicas of the original MLP layers from the dense checkpoint. The router parameters are randomly initialized following a normal distribution with a mean of 0 and a variance of 0.02.
    
    \item \textbf{Continuous Pretraining of MoE}: During both training and inference, two out of eight experts are activated for each token, resulting in approximately 30B activated parameters. To prevent training collapse, additional load balancing loss \cite{fedus2022switch} and max z-loss \cite{chowdhery2023palm, zoph2022st} are applied to the final training objective. The auxiliary loss and max z-loss are multiplied by 0.001 and 0.01, respectively, to ensure a balanced distribution of tokens assigned to different experts and a stable training trajectory.
\end{itemize}

By following this structured approach, EfficientScale enables efficient training of large-scale models through systematic preparation, scaling up, and scaling out. This methodology leverages pre-trained smaller models to reduce data and computational requirements while ensuring efficient knowledge transfer and continuous learning. The result is a highly optimized MoE model capable of performing complex tasks with enhanced efficiency and performance.

\section{Experiemnts}
\subsection{Datasets Description}
We constructed a bilingual pretraining dataset of 4TB tokens in both Chinese and English. This dataset includes webpages, arXiv papers, encyclopedic data, books, codes, and QA pairs. It covers a wide range of high-quality open-source pretraining data such as \href{https://huggingface.co/datasets/togethercomputer/RedPajama-Data-V2}{RedPajama-Data-V2}, \href{https://huggingface.co/datasets/tiiuae/falcon-refinedweb}{falcon-refinedweb}, \href{https://huggingface.co/datasets/allenai/c4}{C4}, \href{https://huggingface.co/datasets/EleutherAI/pile}{Pile}, \href{https://data.baai.ac.cn/details/WuDaoCorporaText}{WuDaoCorporaText}, \href{https://huggingface.co/datasets/CASIA-LM/ChineseWebText}{ChineseWebText}, etc. The above open-source data underwent language filtering to retain only Chinese and English texts, heuristic refinement to remove low-quality content, deduplication to maintain uniqueness, domain-specific filtering for relevance, data quality checks, removal of toxic and explicit content, and finally, data mixing in specified proportions.


\subsection{Experimental Setups and Results}


\subsubsection{Scale-up Validation}
\label{sec:cmp_init}


For the scale-up experiment, we used a 1.3B Aquila2~\footnote{https://github.com/FlagAI-Open/Aquila2} architecture model as the baseline. This model was scaled up to a 7B model using two different methods: FPI and AKI. Additionally, a 7B model was trained from scratch to serve as a control. All three 7B models were trained using the same hyperparameters and on the same dataset for a specified number of steps. We use $\mathcal{M}(24, 2048)$ to denote the 1.3B model with 24 layers and 2048 hidden dimensions and use $\mathcal{M}(32, 4096)$ to denote the 7B model.
We first calculated the validation loss of models with different initializations. The results are shown in Table~\ref{tab:init-losses}. We check the loss of an intermediate model $\mathcal{M}(24, 4096)$ without doing depth growth. We got exactly the same loss as the original model using FPI. Moreover, we found that with interpolation, both FPI and AKI have lower initial losses.

\begin{figure}
    \centering
    \begin{minipage}[t]{0.48\linewidth}
        \centering
        \includegraphics[width=\linewidth]{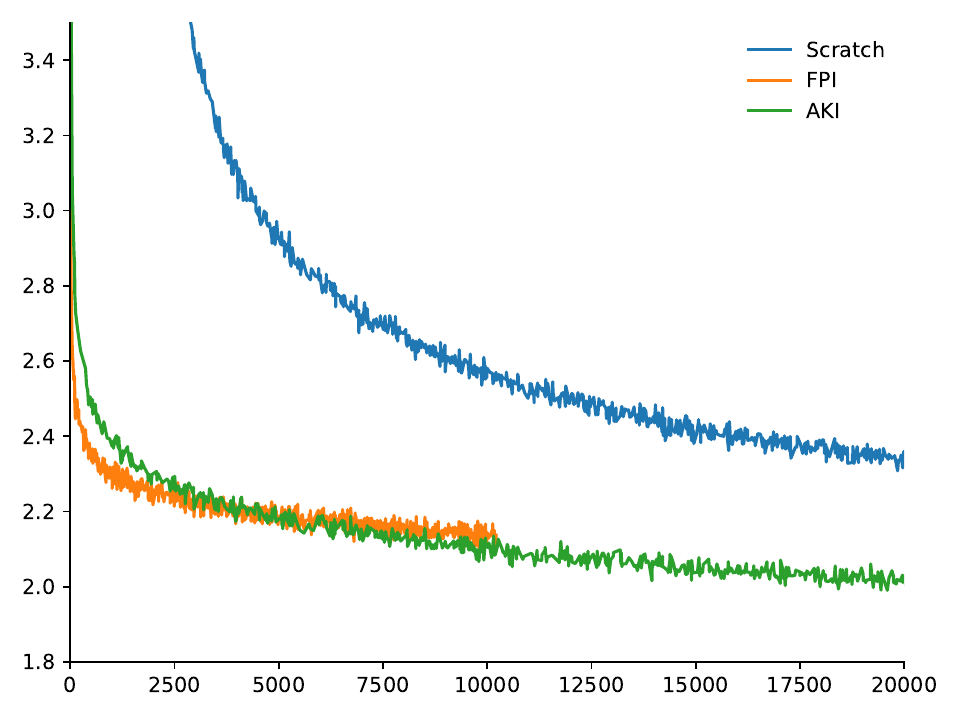}
        \caption{Comparison between the convergence of FPI and AKI methods.}
        \label{fig:fpi-aki-cmp}
    \end{minipage}
    \hfill
    \begin{minipage}[t]{0.48\linewidth}
        \centering
        \includegraphics[width=\linewidth]{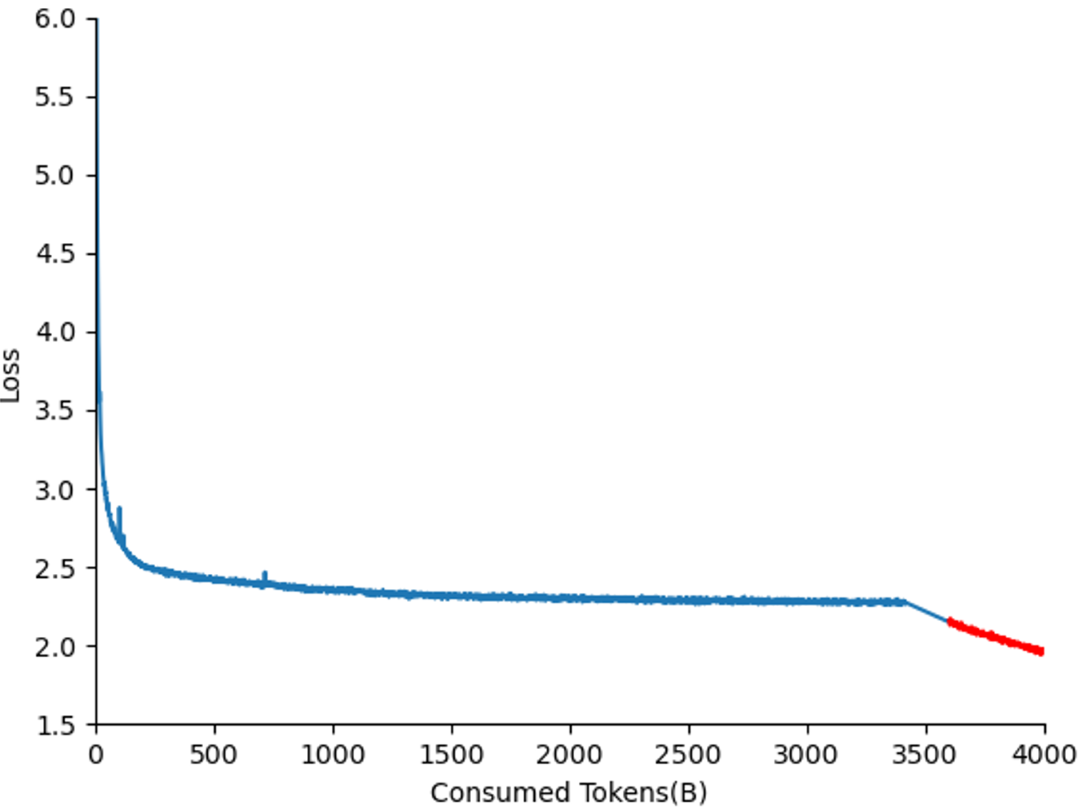}
        \caption{Training loss of AquilaMoE.}
        \label{fig:18btrainloss}
    \end{minipage}
\end{figure}

The loss convergence for the training process is shown in Figure~\ref{fig:fpi-aki-cmp}. The experimental results indicate that the 7B models initialized using the FPI and AKI methods exhibited significantly lower loss values compared to the 7B model trained from scratch. Furthermore, these models converged at a notably faster rate. Consistent with findings in the paper~\cite{chen-etal-2022-bert2bert}, our results also demonstrate that the AKI method surpasses FPI in performance after a certain number of steps.

\begin{table}[h]
\centering
\caption{Validation losses of different initialization methods.}
\label{tab:init-losses}

\begin{tabular}{@{}l|cc@{}}
\toprule
Method                             & \multicolumn{1}{l|}{$\mathcal{M}(24, 4096)$} & \multicolumn{1}{l}{$\mathcal{M}(32, 4096)$} \\ \midrule
$\mathcal{M}(24, 2048)$ (Original) & \multicolumn{2}{c}{$\it 2.97$}                                                                 \\ \midrule
Random                             & \multicolumn{1}{c|}{-}                       & $12.22$                                     \\
FPI (Stacking)                     & \multicolumn{1}{c|}{$\it 2.97$}                  & $4.30$                                      \\
FPI (Interpolation)                & \multicolumn{1}{c|}{$\it 2.97$}                  & $\bf 3.31$                                      \\
AKI (Stacking)                     & \multicolumn{1}{c|}{-}                       & $9.56$                                      \\
AKI-Pro (Interpolation)            & \multicolumn{1}{c|}{-}                       & $\bf 7.81$                                      \\ \bottomrule
\end{tabular}

\end{table}

\begin{table}[h]
\centering
\caption{Validation losses of the AquilaDense-16B initializations. $\mathcal{M}(32, 4096)$ is 7B. $\mathcal{M}(40, 5120)$ is 13B. $\mathcal{M}(32, 5120)$ and $\mathcal{M}(32, 8192)$ are for checking loss before depth growth.}
\label{tab:init-losses-16b}

\begin{tabular}{@{}l|ccc@{}}
\toprule
Method                             & \multicolumn{1}{l|}{$\mathcal{M}(32, 8192)$} & \multicolumn{1}{l|}{$\mathcal{M}(32, 5120)$} & \multicolumn{1}{l}{$\mathcal{M}(40, 5120)$} \\ \midrule
$\mathcal{M}(32, 4096)$ (Original) & \multicolumn{3}{c}{$\it 1.85$}                                                                                                                \\ \midrule
FPI                                & \multicolumn{1}{c|}{$\it 1.85$}                  & \multicolumn{1}{c|}{$1.96$}                  & $2.24$                                      \\
AKI-Pro                            & \multicolumn{1}{c|}{-}                       & \multicolumn{1}{c|}{-}                       & $7.90$                                      \\ \bottomrule
\end{tabular}

\end{table}

\subsubsection{Scale-out Validation}

For the scale-out validation experiment, we trained a 1.8B model from scratch with a training data volume of 3.6T tokens. These models were then scaled out to 8*1.8B configurations, followed by continuous pretraining with an additional 400B tokens. The respective model configurations and training hyperparameters are detailed in Table~\ref{table:models}. We analyzed the loss convergence on the training set with the results depicted in Figure~\ref{fig:18btrainloss}.












\begin{table}[t]  
\centering  
\caption{Model configurations and training parameters for different models.}

\resizebox{\textwidth}{!}{
\begin{tabular}{c|ccccc} 
\toprule  
 & \textbf{1.8B} & \textbf{8*1.8B} & \textbf{7B} (AquilaDense-7B) & \textbf{16B} (AquilaDense-16B) & \textbf{8*16B} (AquilaMoE) \\ 
\midrule  
\textbf{Context Length} & 2048 & 2048 & 4096 & 4096 & 4096 \\   
  
\textbf{QKV Bias} & yes & yes & yes & yes & yes \\   
  
\textbf{Layers} & 24 & 24 & 32 & 40 & 40 \\   
  
\textbf{Hidden Dim} & 2048 & 2048 & 4096 & 5120 & 5120 \\   
  
\textbf{Intermediate Dim} & 5504 & 5504 & 14336 & 20480 & 20480 \\   
  
\textbf{Heads Num} & 32 & 32 & 32 & 40 & 40 \\   
  
\textbf{KV Group} & 32 & 32 & 32 & 8 & 8 \\   
  
\textbf{Trained Tokens (B)} & 3600 & 400 & 3600 & 1200 & 545 \\   
  
\textbf{LR} & 1.20e-3 & 2.20e-4 & 1.20e-3 & 4.00e-4 & 1.50e-4 \\   
  
\textbf{Batch Size} & 12M & 12M & 12M & 12M & 24M \\   
\bottomrule  
\end{tabular}
}

\label{table:models}  
\end{table}

Based on the results of the aforementioned validation experiments, we verified the effectiveness of both scale-up and scale-out approaches on smaller-sized models. Specifically, we trained a model from scratch with a size of 7B, and pre-trained it on 3.6T tokens, resulting in AquilaDense-7B. Subsequently, we scaled it up to a model with a size of 16B and further trained it on 1.2T tokens, yielding AquilaDense-16B. Finally, we scaled it out to 8*16B and trained it on 545B tokens, ultimately obtaining AquilaMoE. The configurations and training parameters of the models are presented in Table~\ref{table:models}.

\section{Model Evaluation}
\subsection{Evaluation of Foundation Models}


\begin{table}[ht]
\centering
\caption{Overall evaluation results of AquilaDense and AquilaMoE(AquilaMoE-8*16B)}
\begin{tabular}{c|ccc}
\toprule
Model  & AquilaDense-7B & AquilaDense-16B & AquilaMoE \\ 
\midrule
ARC-c-ppl       & 37.63 & 38.31 & 43.05 \\
ARC-e-ppl       & 56.08 & 52.2  & 65.61 \\
Hellaswag-ppl   & 67.49 & 71.62 & 73.94 \\
GSM8K-gen       & 7.81  & 28.51 & 54.51 \\
HumanEval-gen   & 14.02 & 29.88 & 15.85 \\
MMLU-ppl        & 46.47 & 57.11 & 61    \\
Winograd-ppl    & 50.53 & 54.04 & 55.4  \\
MATH-gen        & 1.32  & 4.24  & 10.4  \\
MBPP-gen        & 15.6  & 36.4  & 37.2  \\
DROP-gen        & 4.35  & 33.35 & 37.62 \\
AGI Eval-gen     & 14.47 & 18.57 & 13.69     \\
BBH-gen         & 34.51 & 41.45 & 46.04 \\
NQ-gen          & 8.61  & 9.94  & 10.78 \\
PIQA-ppl        & 76.71 & 79.22 & 80.3  \\
\bottomrule
\end{tabular}
\label{tab:moe-main}
\end{table}

Following OpenCompass\footnote{https://github.com/open-compass}, in the evaluation process, we use two types of evaluation methods: discriminant analysis evaluation and generative evaluation. Discriminant analysis evaluation means combining the question with candidate answers, calculating the perplexity of all combinations, and selecting the answer with the lowest perplexity as the model's final output. Generative evaluation uses the question as the model's original input and leaves the answer area blank for the model to complete subsequently.

The performance of AquilaDense-7B, AquilaDense-16B, and AquilaMoE(8*16B) models are presented in Table~\ref{tab:moe-main}. The indicators ending in ``ppl'' represent discriminant analysis evaluation, while those ending in ``gen'' represent generative evaluation.

Generally, as the model size increases, the scores tend to improve. For instance, AquilaDense-7B scores 7.81 on GSM8K-gen, while AquilaDense-16B scores 28.51. A similar trend also is observed in most other tasks. The AquilaMoE models show improved performance in most tasks over AquilaDense-16B. For example, in the ARC-c-ppl task, AquilaMoE scored 43.05 compared to 38.31 for AquilaDense-16B. These findings highlight the benefits of both scaling up model parameters and implementing MoE architectures in improving model performance.

\subsection{Evaluation of Fine-tuned Models}
Table~\ref{tab:moe-main-sft} presents the overall results of AquilaMoE-8*16B after fine-tuning across various benchmark datasets. The performance is measured using generative evaluation, and the results are expressed as percentages.

\begin{table}[ht]
\centering
\caption{Overall results of AquilaMoE after fine-tuning.}
\begin{tabular}{cc}
\toprule
Model  &AquilaMoE-8*16B-SFT \\ \midrule
ARC-c  &82.03  \\
ARC-e  &87.3  \\          
Hellaswag  &75.08 \\  
GSM8K  &71.27 \\  
NQ &21.39 \\
TriviaQA &65.33 \\
AGI Eval & 13.61 \\
Math  & 13.26 \\  
HumanEval &44.51 \\
PIQA &81.72 \\
OBQA &75.2 \\
DROP &62.32 \\
BoolQ &85.02 \\
GPQA &25.76 \\
C-Eval & 57.99\\
MMLU & 61.51 \\
CMMLU &57.63 \\
Winogrande & 57.54 \\
\bottomrule
\end{tabular}
\label{tab:moe-main-sft}
\end{table}


\subsection{Comparsion of Computational Efficiency}

We present the details of the training process for both scale-up + scale-out and from-scratch approaches in Table~\ref{tab:detail}. The table lists the number of devices in the cluster, the GFLOPS per device, the model parameters size, the number of trained tokens, the actual number of training tokens per day, the actual training time, and the actual training GFLOPS for each phase.

\begin{table}[t]
\centering
\caption{Training details for scale-up and scale-out and from-scratch approaches, note that for preparation phase different chip is used.}

\resizebox{\textwidth}{!}{
\begin{tabular}{@{}c|ccccc@{}}
\toprule
\textbf{Approach/Phase} & \textbf{Devices} & \textbf{GFLOPS/Device} & \textbf{Model Size (B)} & \textbf{Trained Tokens (B)} & \textbf{Training Tokens/Day (B)} \\ \midrule
 Preparation Phase & 480 & 989.5 & 7 & 3600 & 279 \\
 Scale-Up Phase & 1024 & 240 & 16 & 1200 & 70 \\
 Scale-Out Phase & 1024 & 240 & 32 & 545 & 25 \\ \midrule
From Scratch & 1024 & 240 & 32 & 5345 & 25 \\ \bottomrule
\end{tabular}
}

\label{tab:detail}
\end{table}

The time savings factor is calculated by comparing the total training time of the from-scratch approach to the total training time of the scale-up and scale-out approach. The formula is:

\[
\text{Time Savings Factor} = \frac{\frac{\sum_{i=1}^{n} N_{\text{tokens}, i}}{R_{\text{tokens/day, from scratch}}}}{\sum_{i=1}^{n} \frac{N_{\text{tokens}, i}}{R_{\text{tokens/day}, i}}}
\]

Given the data:
\[
\text{Time Savings Factor} = \frac{\frac{3600 + 1200 + 545}{25}}{\frac{3600}{279} + \frac{1200}{70} + \frac{545}{25}} = \frac{213.80}{51.84} \approx 4.12
\]

The computational power savings factor is calculated by comparing the total GFLOPS-days of the from-scratch approach to the total GFLOPS-days of the scale-up and scale-out approach. The formula is:

\[
\text{Computational Power Savings Factor} = \frac{\frac{\sum_{i=1}^{n} N_{\text{tokens}, i} \times \text{GFLOPS}_{\text{from scratch}}}{R_{\text{tokens/day, from scratch}}}}{\sum_{i=1}^{n} \frac{N_{\text{tokens}, i} \times \text{GFLOPS}_i}{R_{\text{tokens/day}, i}}}
\]

Given the data:
\[
\text{GFLOPS}_{\text{preparation}} = 480 \times 989.5 = 475,360
\]
\[
\text{GFLOPS}_{\text{scale-up}} = 1024 \times 240 = 245,760
\]
\[
\text{GFLOPS}_{\text{scale-out}} = 1024 \times 240 = 245,760
\]
\[
\text{GFLOPS}_{\text{from scratch}} = 1024 \times 240 = 245,760
\]

The computational power savings factor is:
\[
\text{Computational Power Savings Factor} = \frac{\frac{5345 \times 245,760}{25}}{\frac{3600 \times 475,360}{279} + \frac{1200 \times 245,760}{70} + \frac{545 \times 245,760}{25}} = \frac{52,592,640}{15,705,343} \approx 3.35
\]

The method proposed in this paper significantly reduces both the computational power and the time required for training. By employing a scale-up and scale-out approach, we achieved a computational power savings factor of approximately 3.35 and a time savings factor of approximately 4.12. 

Additionally, if we start with a pre-trained smaller model, the computational power and time required for the preparation phase can be further reduced. This approach not only accelerates the training process but also lowers the overall computational costs.

In summary, the proposed training methodology offers substantial improvements in efficiency. The combined scale-up and scale-out approach, along with the potential use of pre-trained models, represents a significant advancement in the optimization of training large-scale models.

\section{Conclusion and Future Work}

We present AquilaMoE, a bilingual 8*16B mixture of experts (MoE) language model developed using the EfficientScale training method. EfficientScale optimizes performance while significantly reducing data requirements through a two-stage approach: Scale-Up and Scale-Out. Our contributions are as follows: 1) An effective training methodology that achieves knowledge transfer and continuous pretraining with significantly reduced data and computational needs; 2) Innovative initialization strategies, such as Functional Progressive Initialization (FPI) and Approximate Knowledge Integration (AKI), which demonstrate substantial loss retention and reduction during continual pre-training; 3) Successful training of 16B and 8*16B AquilaMoE models using these initialization strategies, enhancing performance and training efficiency. Future work involves exploring the scalability of larger MoE models, investigating cross-linguistic knowledge transfer, developing new optimization techniques to further reduce training time and costs, fine-tuning for specific application domains, and ensuring the robustness and generalization of MoE models across diverse datasets and real-world applications.

\section*{Authorship}
\textbf{Language Foundation Model \& Software Team, BAAI}: Bo-Wen Zhang, Liangdong Wang, Jijie Li, Shuhao Gu, Mengdi Zhao, Xinya Wu, Guang Liu~(\textbf{Project lead})\footnote{The correspinding author, contact \url{liuguang@baai.ac.cn}.}\\

\textbf{Data Research Team, BAAI}: Chengwei Wu, Hanyu Zhao, Li Du, Yiming Ju, Quanyue Ma\\

\textbf{AI Framework Research and Development Team, BAAI}: Yulong Ao (\textbf{Infrastructure lead}), Yingli Zhao, Songhe Zhu,  Zhou Cao, Dong Liang, Yonghua Lin\\

\textbf{School of Computer Science, Peking University}: Ye Yuan\footnote{Responsible for the full design and implementation of the Scale-Up strategy. Main work done during his internship at BAAI.}, Ming Zhang\\

\textbf{MetaX-Tech}: Shunfei Wang, Yanxin Zhou, Min Ye, Xuekai Chen, Xinyang Yu, Xiangjun Huang, Jian Yang

\bibliographystyle{unsrt}  
\bibliography{references}

\end{document}

%% file: sections/scale-up.tex
\subsection{Scale-Up Phase}
The Scale-Up phase involves two critical steps: initializing the weights of a larger dense model using the smaller model and performing continuous pretraining to ensure effective knowledge transfer and model enhancement. We use the bert2BERT\cite{chen-etal-2022-bert2bert} method to initialize the large model and propose the AKI-Pro method, improving bert2BERT-AKI from depth expansion and group query attention.

\subsubsection{Weight Initialization Strategies}
The weights of the small dense model are used to initialize a larger dense model. There are two strategies proposed in bert2BERT\cite{chen-etal-2022-bert2bert}: Function Preserving Initialization(FPI) and Advanced Knowledge Initialization(AKI). Both the original and our experiments in Section \ref{sec:cmp_init} show that AKI performs better. Besides, recent research\cite{pan2024preparing} shows that it is better to use interpolation instead of stacking when expanding the depth, which is more stable for continuous training. Moreover, the original AKI method is not suitable for Group Query Attention (GQA), so we modify the transformation of the weights in attention blocks to fit GQA. Finally, we have AKI-Pro as our initialization method. Below we will introduce the three initialization methods, starting with a review of the first two approaches in bert2BERT, followed by our improvements.

\begin{figure}[t]
    \centering
    \includegraphics[width=0.95\linewidth]{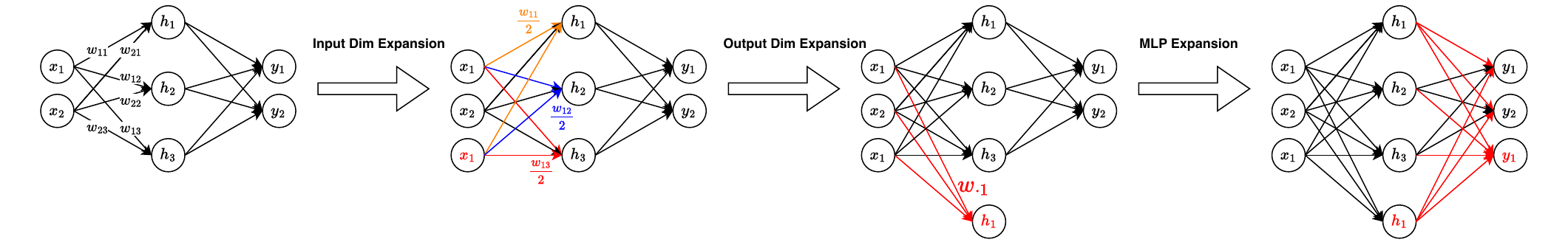}
    \caption{An example of FPI on an MLP layer.}
    \label{fig:fpi-example}
\end{figure}

\textbf{Function Preserving Initialization (FPI)}: This strategy is firstly proposed in Net2Net\cite{net2net} to expand the intermediate dim of an MLP layer. Bert2BERT\cite{chen-etal-2022-bert2bert} enhances the Net2Net method to FPI, which enables it to expand the hidden dims(i.e. input and output dims). It is applied in training language models in bert2BERT and can expand the width of a smaller model to a larger model, getting the same output with the same input. With the FPI, the larger model can get the transferred knowledge from the smaller model. The basic idea behind FPI is that when expanding the dims, it makes both the input and output tensor concatenate a copy of the smaller tensor, as illustrated in Figure~\ref{fig:fpi-example}. For an MLP layer with two linear mappings in the example: $\boldsymbol{y}=\boldsymbol{U}^{\top}\boldsymbol{W}^{\top}\boldsymbol{x}$, the input and output dims are 2, and the intermediate dim is 3. Suppose we want to expand this block to that with 3 as input and output dims, and 4 as intermediate size, then there are three steps. \textbf{(1) Input Dim Expansion} FPI copies the input neurons from left to right and splits the corresponding weights to the new input neurons. \textbf{(2) Output Dim Expansion} For the expansion of the output in the upsampling linear weights, FPI also makes the new hidden neurons copy from the original ones. \textbf{(3) MLP Expansion} Expand the downsampling linear weights the same as the upsampling weights, and finally, the new output neurons of this MLP layers are also the copy from the smaller ones, which makes the block can be stacked as layers. The weights $\boldsymbol{W'}=\textbf{FPI}\left(\boldsymbol{W}\right)$ are transformed as follows:
\begin{equation}
    \begin{split}
    \boldsymbol{w'}_{1,*} &= \boldsymbol{w'}_{3,*} = \frac{\boldsymbol{w}_{1,*}}{2} \\
    \boldsymbol{w'}_{*,4} &= \boldsymbol{w'}_{*,1}
    \end{split}
\end{equation}
Most modules of a transformer block can be transformed the same as an MLP layer, including embedding layers and QKV projections. For the MHA module, each attention head should be seen as a neuron, and then the head number can be expanded as before. Notably, the output of the LN modules will not be the same when the new dimension is not an integer multiple of the old one, but this will not hurt a lot on the final loss.

\textbf{Advanced Knowledge Initialization (AKI)}: As shown in both Net2Net\cite{net2net} and bert2BERT\cite{chen-etal-2022-bert2bert}, the symmetry from the FPI will hinder the model convergence. Specifically, if we have a linear layer $y=w_1x+w_2x$, where $x, y\in \mathbb{R}$, and $w_1=w_2$ when initializing the weights, the gradient and the value of these two weights will always be the same, which makes the effective number of parameters for this linear layer only 1. So AKI is proposed to break the symmetry with expanding width based on not only the weights of the same layer but also the upper layer in the smaller model. Take a model with two MLP blocks as an example:
\begin{equation}
    \begin{split}
        & \boldsymbol{y_1} = \boldsymbol{U^{(1)\top}}\boldsymbol{W^{(1)\top}}\boldsymbol{x}, \boldsymbol{y_2} = \boldsymbol{U^{(2)\top}}\boldsymbol{W^{(2)\top}}\boldsymbol{y_1}, \boldsymbol{x}, \boldsymbol{y_1}, \boldsymbol{y_2} \in \mathbb{R}^2 \\
        & \boldsymbol{W^{(1,2)}} \in \mathbb{R}^{2\times 3}, \boldsymbol{U^{(1,2)}} \in \mathbb{R}^{3\times 2} \\
    \end{split}
\end{equation}
FPI expands $\boldsymbol{W^1}$ as $\textbf{FPI}\left(\boldsymbol{W^{(1)}}\right)=\left[\boldsymbol{w_1'^{(1)};w_2'^{(1)};w_3'^{(1)};w_1'^{(1)}}\right]$, while AKI uses the output expansion of next layer: $\textbf{AKI}\left(\boldsymbol{W^{(1)}}\right)=\left[\boldsymbol{w_1'^{(1)};w_2'^{(1)};w_3'^{(1)};w_1'^{(2)}}\right]$. Inspired by the observation that neighboring layers have similar functions, AKI breaks the symmetry and keep the knowledge from the smaller models. Moreover, FPI can't expand the depth, so bert2BERT uses the stacking method to expand the model depth proposed by StackBERT~\cite{stacking}.

\begin{figure}
    \centering
    \includegraphics[width=0.6\linewidth]{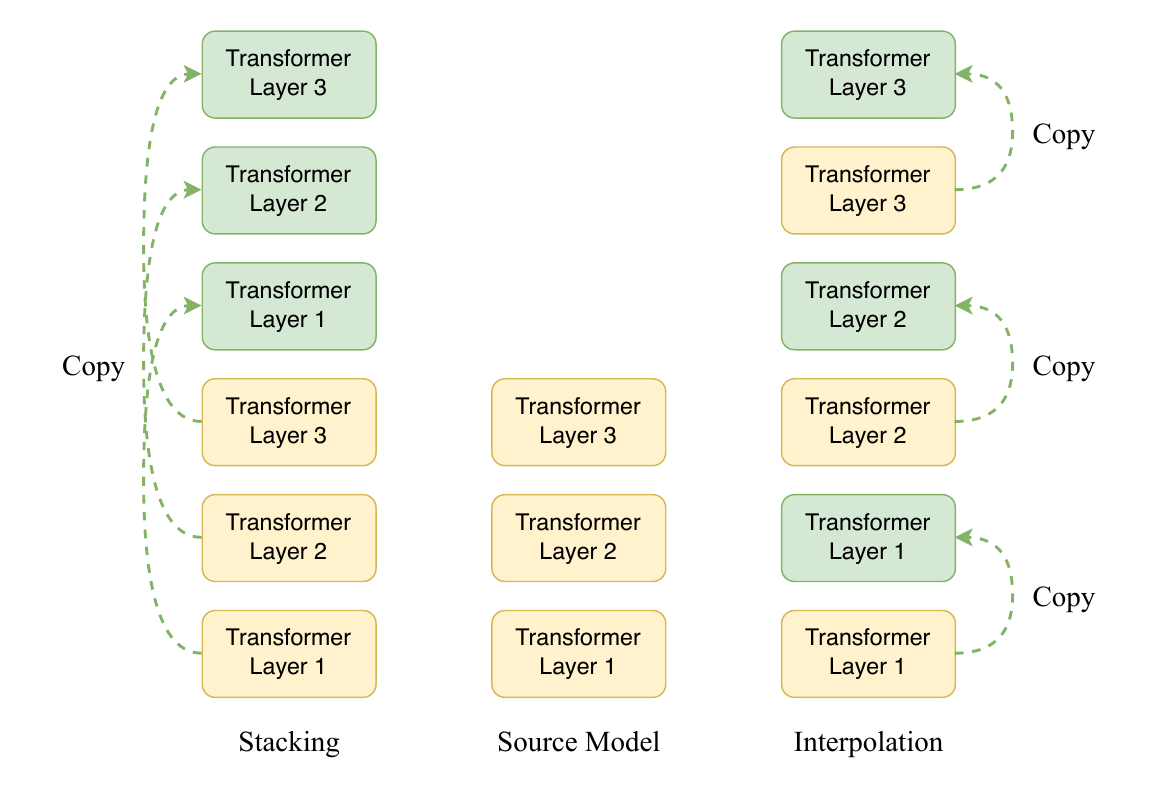}
    \caption{Comparison of different growing methods: stacking and interpolation.}
    \label{fig:stack-inter}
\end{figure}

\textbf{AKI-Pro}: Our proposed improvement on AKI further refines weight initialization from two aspects: depth growing method and GQA compatibility. 
\begin{itemize}
    \item \textbf{Depth Growing Method:} We use interpolation in the depth growth to make the continuous training more stable, following the recent research \cite{pan2024preparing}. The stacking method just copies the layers of the source model to the top. For the source model with $L_1$ layers: $\left\{W_l | l\in [0, L_1)\right\}$ and target model with $L_2$ layers: $\left\{W'_l | l\in [0, L_1)\right\}$, stacking method can be formed as $W'_l=W_{(l \mod L_1)}$. However, the output space of the last layer does not match the input space of the first layer, which can make the continuous training unstable. Based on the observation of similar functionality in neighboring layers, recent research\cite{pan2024preparing} improves this by using interpolation, which can be formulated as below:
    \begin{equation}
        W'_l = \lfloor \frac{l * L_2}{L_1} \rfloor
    \end{equation}
    Figure~\ref{fig:stack-inter} shows an example when $L_1=3, L_2=6$. We show the comparison of validation losses and training curves after the depth growth with different methods in Section~\ref{sec:cmp_init}. 
    \item \textbf{GQA Compatibility:} The original AKI method only supports MHA in transformer models. We adapt AKI for Group Query Attention models. To be specific, under the constraint that the number of groups in the GQA of the source model and the target model are consistent, we expand the output of the attention heads inside each group. Each group can be seen as a separate MHA block with common KV projection weights, and the expansion operator is the same as MHA.
\end{itemize}

\subsubsection{Continuous Pretraining Process}
The scaled-up dense model undergoes continuous pretraining on a substantial amount of tokens. This phase ensures the successful transfer of knowledge and allows the model to acquire additional information from the data, enhancing its overall performance and capability.